%% file: acl2021.tex
\documentclass[11pt,a4paper]{article}
\usepackage[hyperref]{acl2021}
\usepackage{times}
\usepackage{multirow}
\usepackage[ruled,vlined]{algorithm2e}

\usepackage{algpseudocode}
\usepackage{latexsym}
\usepackage{amsmath}
\usepackage{graphicx}
\usepackage{booktabs}
\usepackage{microtype}

\aclfinalcopy 
\def\aclpaperid{395} 

\SetKwRepeat{Do}{do}{while}%

\title{Adapt-and-Distill: Developing Small, Fast and Effective \\ Pretrained Language Models for Domains}
\author{
  Yunzhi Yao$^\dag$\thanks{\; Contribution during internship at Microsoft Research.}, Shaohan Huang$^\ddag$, Wenhui Wang$^\ddag$, Li Dong$^\ddag$, Furu Wei$^\ddag$  \\
  $^\dag$Shandong University, Jinan, China \\
  $^\ddag$Microsoft Research, Beijing, China \\
  \texttt{yyz@mail.sdu.edu.cn} \\
  \texttt{\{shaohanh,wenhui.wang,lidong1,fuwei\}@microsoft.com} \\}

\date{}

\begin{document}
\maketitle
\begin{abstract}
Large pretrained models have achieved great success in many natural language processing tasks. 
However, when they are applied in specific domains, these models suffer from domain shift and bring
challenges in fine-tuning and online serving for latency and capacity constraints.
In this paper, we present a general approach to developing small, fast and effective pretrained models for specific domains. This is achieved by adapting the off-the-shelf general pretrained models and performing task-agnostic knowledge distillation in target domains.
Specifically, we propose domain-specific vocabulary expansion in the adaptation stage and employ corpus level occurrence probability to choose the size of incremental vocabulary automatically.
Then we systematically explore different strategies to compress the large pretrained models for specific domains.
%
We conduct our experiments in the biomedical and computer science domain. The experimental results demonstrate that our approach achieves better performance over the $\text{BERT}_{\text{BASE}}$ model in domain-specific tasks while 3.3× smaller and 5.1× faster than $\text{BERT}_{\text{BASE}}$.
The code and pretrained models are available at \url{https://aka.ms/adalm}.

\end{abstract}

\input{sec/Introduction}

\input{sec/Relatedwork}

\input{sec/Method}

\input{sec/ExperimentalDetails}

\input{sec/Results}

\input{sec/Analysis}

\input{sec/Conclusion}

\input{acl2021.bbl}
\bibliographystyle{acl_natbib}
\clearpage
\input{sec/Appendix}

\end{document}

%% file: sec/Introduction.tex
\section{Introduction}

Pre-trained language models, such as GPT~\cite{radford2018improving}, BERT~\cite{devlin-etal-2019-bert}, RoBERTa~\cite{liu2019roberta} and UniLM~\cite{dong2019unified} have achieved impressive success in many natural language processing tasks. These models usually have hundreds of millions of parameters. They are pre-trained on a large corpus of general domain and fine-tuned on target domain tasks. However, it is not optimal to deploy these models directly to edge devices in specific domains. First, heavy model size and high latency makes it difficult to deploy on resource-limited edge devices such as mobile phone. Second, directly fine-tuning a general pre-trained model on a domain-specific task may not be optimal when the target domain varies substantially from the general domain. Thirdly,  many specialized domains contain their own specific terms, which are not included in pre-trained language model vocabulary.

\begin{figure}[t]
   \centering
   \includegraphics[scale=0.41]{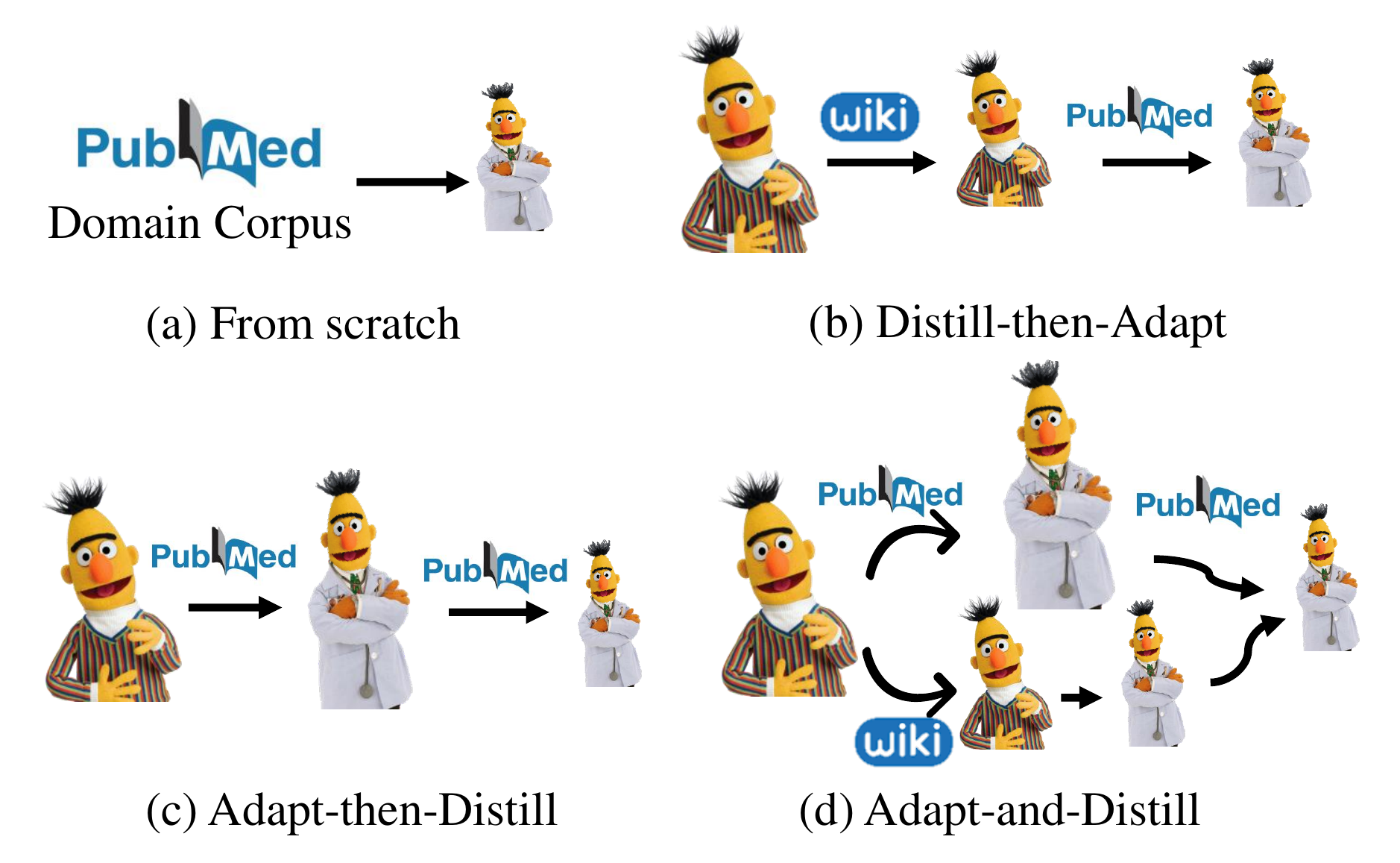}
   \caption{The four alternatives when distilling BERT into specific domains. All strategies are task-agnostic.}
   \label{fig:overview}
\end{figure}

In this paper, we introduce AdaLM, a framework that aims to develop small, fast and effective pre-trained language models for specific domains. To address domain shift problem, recent studies~\cite{lee2020biobert,dontstoppretraining2020} conduct continual pre-training to adapt a general domain pre-trained model to specific domains. 
However, specific domains contain many common in-domain terms, which may be divided into bite-sized pieces (e.g., \textit{lymphoma} is tokenized into [\textit{l, \#\#ym, \#\#ph, \#\#oma}]). 
\citeauthor{2020Domain}(\citeyear{2020Domain}) mentions that domain-specific vocabularies play a vital role in domain adaptation of pre-trained models. Specifically, we  propose a domain-specific vocabulary expansion in the adaptation stage, which augments in-domain terms or subword units automatically given in-domain text. Also, it is critical to decide the size of incremental vocabulary. Motivated by subword regularization~\cite{kudo2018subword}, AdaLM introduces a corpus occurrence probability as a metric to optimize the size of incremental vocabulary automatically.

We systematically explore different strategies to compress general BERT models to specific domains (Figure~\ref{fig:overview}): (a) From scratch: pre-training domain-specific small model from scratch with domain corpus; (b) Distill-then-Adapt: first distilling large model into small model, then adapting it into a specific domain; (c) Adapt-then-Distill: first adapting BERT into a specific domain, then distilling model into small size; (d) Adapt-and-Distill: adapting both the large and small models, then distilling with these two models initializing the teacher and student models respectively.


We conduct experiments in both biomedical and computer science domain and fine-tune the domain-specific small models on different downstream tasks. Experiments demonstrate that Adapt-and-Distill achieves state-of-the-art results for domain-specific tasks. Specifically, the 6-layer model of 384 hidden dimensions outperforms the $\text{BERT}_{\text{BASE}}$ model while 3.3× smaller and 5.1× faster than $\text{BERT}_{\text{BASE}}$.


%% file: sec/Relatedwork.tex
\section{Related Work}

\paragraph{Domain adaptation of pre-trained model}
Most previous work on the domain-adaptation of pre-trained models targets large models. 
\citeauthor{lee2020biobert}~(\citeyear{lee2020biobert}) conduct continual pre-training to adapt the BERT model to the biomedical domain using the PubMed abstracts and the PMC full text. \citeauthor{dontstoppretraining2020}~(\citeyear{dontstoppretraining2020}) also employ continual pre-training to adapt pre-trained models into different domains including biomedical, computer science and news. However, many specialized domains contain their own specific words that are not included in pre-trained language model vocabulary. \citeauthor{2020Domain}(\citeyear{2020Domain}) propose a biomedical pre-trained model PubMedBERT, where the vocabulary
was created from scratch and the model is pre-trained from scratch. Furthermore, in many specialized domains, large enough corpora may not be available to support pre-training from scratch. \citet{zhang2020multi} and \citet{tai2020exbert} extend the open-domain vocabulary with top frequent in-domain words to resolve this out-of-vocabulary issue. This approach ignores domain-specific sub-word units (e.g., \textit{blasto-}, \textit{germin-} in biomedical domain). These subword units help generalize domain knowledge and avoid unseen words. 

\paragraph{Task-agnostic knowledge distillation}
In recent years, tremendous progress has been made in model compression~\cite{cheng2017survey}. Knowledge distillation has proven to be a promising way to compress large models while maintaining accuracy~\cite{sanh2019distilbert,jiao2019tinybert,sun2020mobilebert,wang2020minilm}. In this paper, we focus on task-agnostic knowledge distillation approaches, where a distilled small pre-trained model can be directly fine-tuned on downstream tasks. DistilBERT~\cite{sanh2019distilbert} employs the soft label and embedding outputs to supervise the student. TinyBERT~\cite{jiao2019tinybert} and MobileBERT~\cite{sun2020mobilebert} introduce self-attention distributions and hidden states to train the student model. MiniLM~\cite{wang2020minilm} avoids restrictions on the number of student layers and employs the self-attention distributions and value relation of the teacher’s last transformer layer to supervise the student model. Because this method is more flexible, we implement MiniLM to compress large models in this work. 
No previous work systematically explores different strategies to achieve an effective and efficient smaller model in specific domains.

%% file: sec/Method.tex
\section{Methods}

\subsection{Overview}
We systematically explore different strategies to achieve an effective and efficient small model in specific domains. We summarize them into four strategies: from scratch, distill-then-adapt, adapt-then-distill and adapt-and-distill.

\paragraph{Pretrain-from-scratch} 
Domain-specific pretraining from scratch employs a random initialization of a pretrained model and pretrains a small model directly on domain-specific corpus. In this work, we conduct pretraining from scratch on different vocabularies including BERT original vocabulary, from scratch vocabulary, and expanded vocabulary.

\paragraph{Distill-then-adapt} 
These approaches first distill the large general pretrained model which pretrained on Wikipedia and BookCorpus. Then it continues the pretraining process using a domain-specific corpus. In this work, we first distill the BERT model into a small model using task-agnostic knowledge distillation in MiniLM~\cite{wang2020minilm}. Then we initialize the small model with it and conduct continual training with both the BERT original vocabulary and the expanded vocabulary.

\paragraph{Adapt-then-distill} In this work, we select different large models as teacher models such as BERT and large models with different vocabularies. We first adapt these models into domain-specific models and then implement MiniLM to compress them to small models.

\paragraph{Adapt-and-distill}
In the previous part, when doing knowledge distill, we initialized the student model randomly.
In order to get a better domain-specific small model, we try to explore the impact of the initialization of the student model. In this part, we adapt large and small models into specific domains separately, then use these two models to initialize the teacher and student model respectively.

\subsection{Domain Adaptation} 
\label{sec:adalm}
\begin{figure}[t]
   \centering
   \includegraphics[scale=0.5]{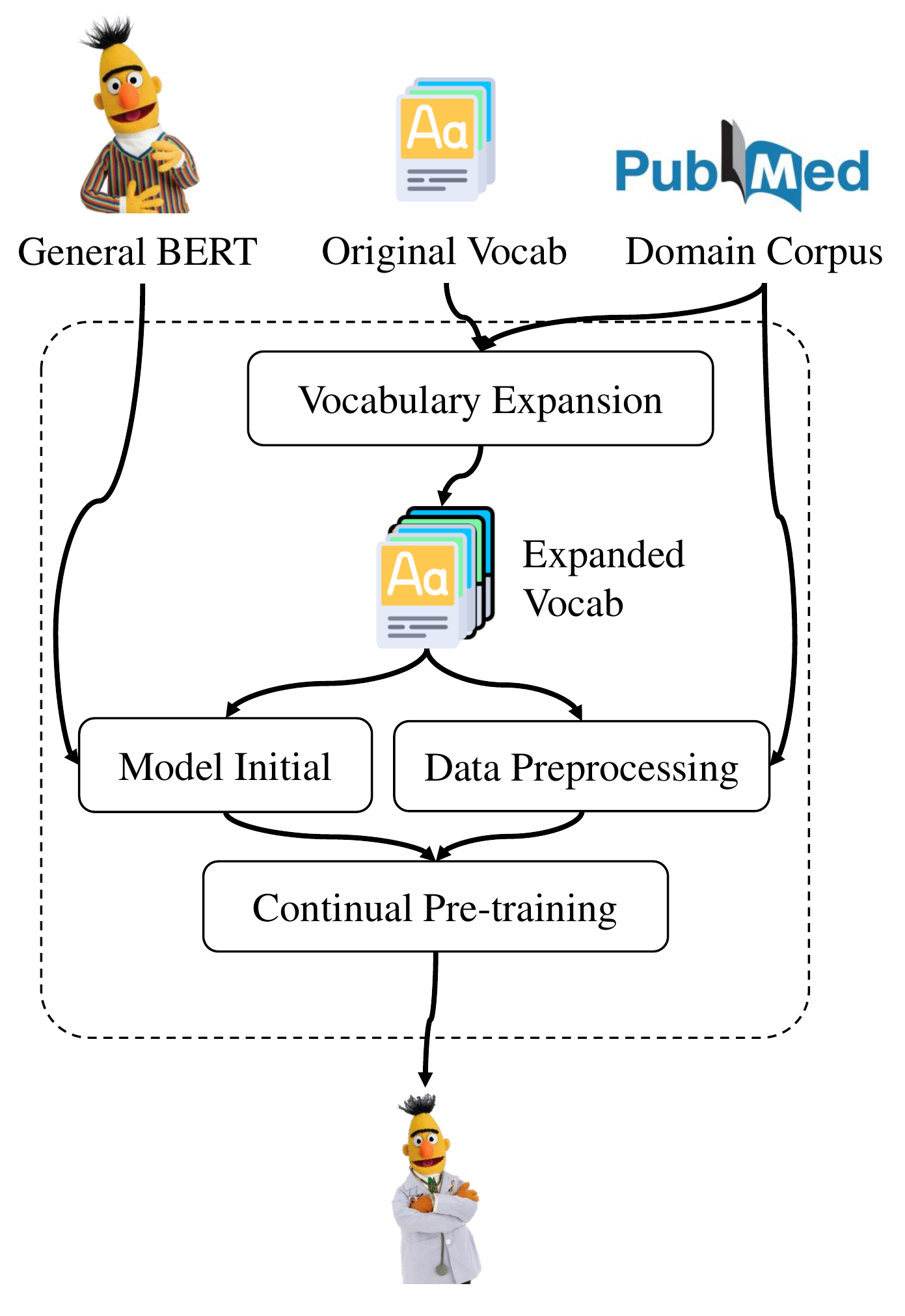}
   \caption{The pipeline of domain adaptation. Here we adapt the BERT model into the biomedical domain with PubMed dataset.}
   \label{fig:pipeline}
\end{figure}

AdaLM contains a simple yet effective domain adaptation framework for a pretrained language model. As shown in Figure~\ref{fig:pipeline}, it takes a general pretrained language model, original vocabulary and a domain specific corpus as input. Through vocabulary expansion and continual pretraining, AdaLM adapts general models into specific domains.

\begin{figure}[t]
   \centering
   \includegraphics[scale=0.43]{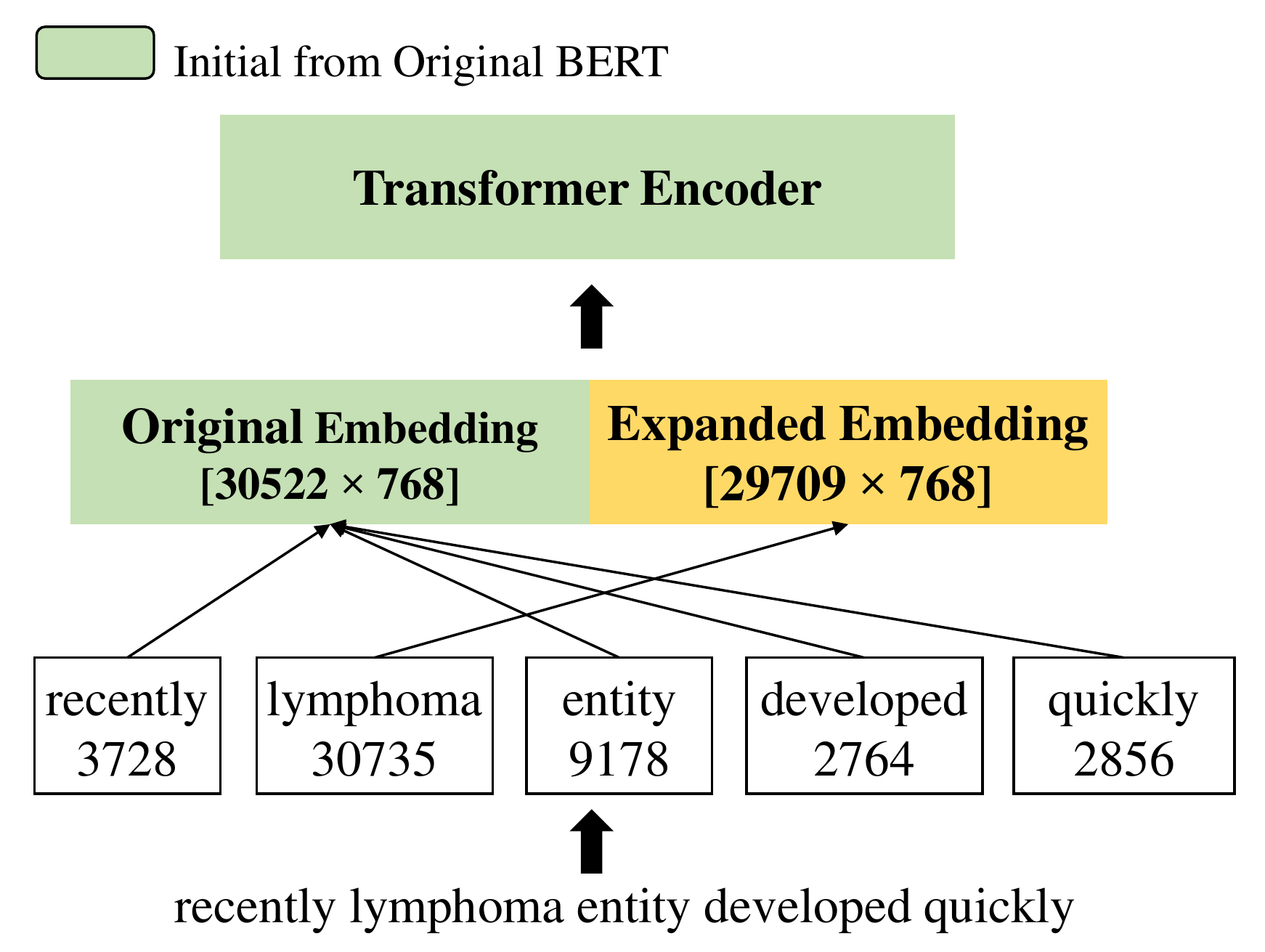}
   \caption{Concatenate original embedding with expanded embedding.}
   \label{fig:model_init}
\end{figure}

The core pipeline of domain adaptation consists of the three steps described below:
\begin{enumerate}
\item Given original vocabulary and a domain-specific corpus, the vocabulary expansion module aims to augment original vocabulary with domain-specific subword units or terms. We augment domain-specific vocabulary from the target domain, while keeping the original BERT vocabulary unchanged. We describe them in more detail in Section~\ref{sec:vocab}.
\item Due to the size of the vocabulary having changed, we cannot initialize our model with BERT directly. As illustrated in Figure~\ref{fig:model_init}, we initialize the original embedding and Transformer encoder with weights from BERT (the green part in Figure~\ref{fig:model_init}). For incremental vocabulary, we first tokenize them into sub-words with the original vocabulary and then use an average pooling of their own sub-words embedding  to initialize. As shown in Figure~\ref{fig:model_init}, the word `\textit{lymphoma}' is not included in BERT vocabulary. We tokenize it into three sub-words (\textit{lym}, \textit{\#\#pho}, \textit{\#\#ma}). The embedding vector of `lymphoma' is initialized  by the average embedding vector of `\textit{lym}', `\textit{\#\#pho}'and `\textit{\#\#ma}'.
\item After model initialization and data preprocessing, we continually pretrain our model with domain-specific corpus using masked language model loss. Following BERT, we randomly replace 15\% of tokens by a special token (e.g., [MASK]) and ask the language model to predict them in continual pretraining.


\end{enumerate}


\subsection{Vocabulary Expansion}\label{sec:vocab}
Vocabulary expansion is the core module of AdaLM. It augments domain-specific terms or subword units to leverage domain knowledge. The size of the incremental vocabulary is a vital parameter for vocabulary expansion. 
Considering that unigram language modeling ~\cite{kudo2018subword} aligns more closely with morphology and avoids problems stemming from BPE’s greedy construction procedure, as proposed in \cite{bostrom-durrett-2020-byte}, we followed \citet{kudo2018subword} and introduced a corpus occurrence probability as a metric to optimize the size of incremental vocabulary automatically. We assume that each subword occurs independently and we assign to each subword in the corpus a probability equal to its frequency in the corpus.
\begin{eqnarray}
      \forall i\,\, x_i \in \mathcal{V},\,\,\,
      \sum_{x_i \in \mathcal{V}} p(x_i) = 1, \label{prob of sentences}
\end{eqnarray}
where $\mathcal{V}$ is a pre-determined vocabulary. The probability of a subword sequence $\mathbf{x} = (x_1,\ldots,x_M)$ can be computed by the product of the subword appearance probabilities $p(x_i)$. 
We convert it to logarithmic form:
\begin{eqnarray}
      P(\mathbf{x}) = \sum_{i=1}^{M} log(p(x_i)), 
\end{eqnarray}

Given a domain-specific corpus $D$, the occurrence probability of corpus $D$ is formulated as:
\begin{eqnarray}
      P(D) = \sum_{\mathbf{x}}^{|D|} log(P(\mathbf{x})),
\end{eqnarray}
where $\mathbf{x}$ represents tokenized sentence in corpus $D$.

\begin{figure}[h] 
   \centering
   \includegraphics[scale=0.52]{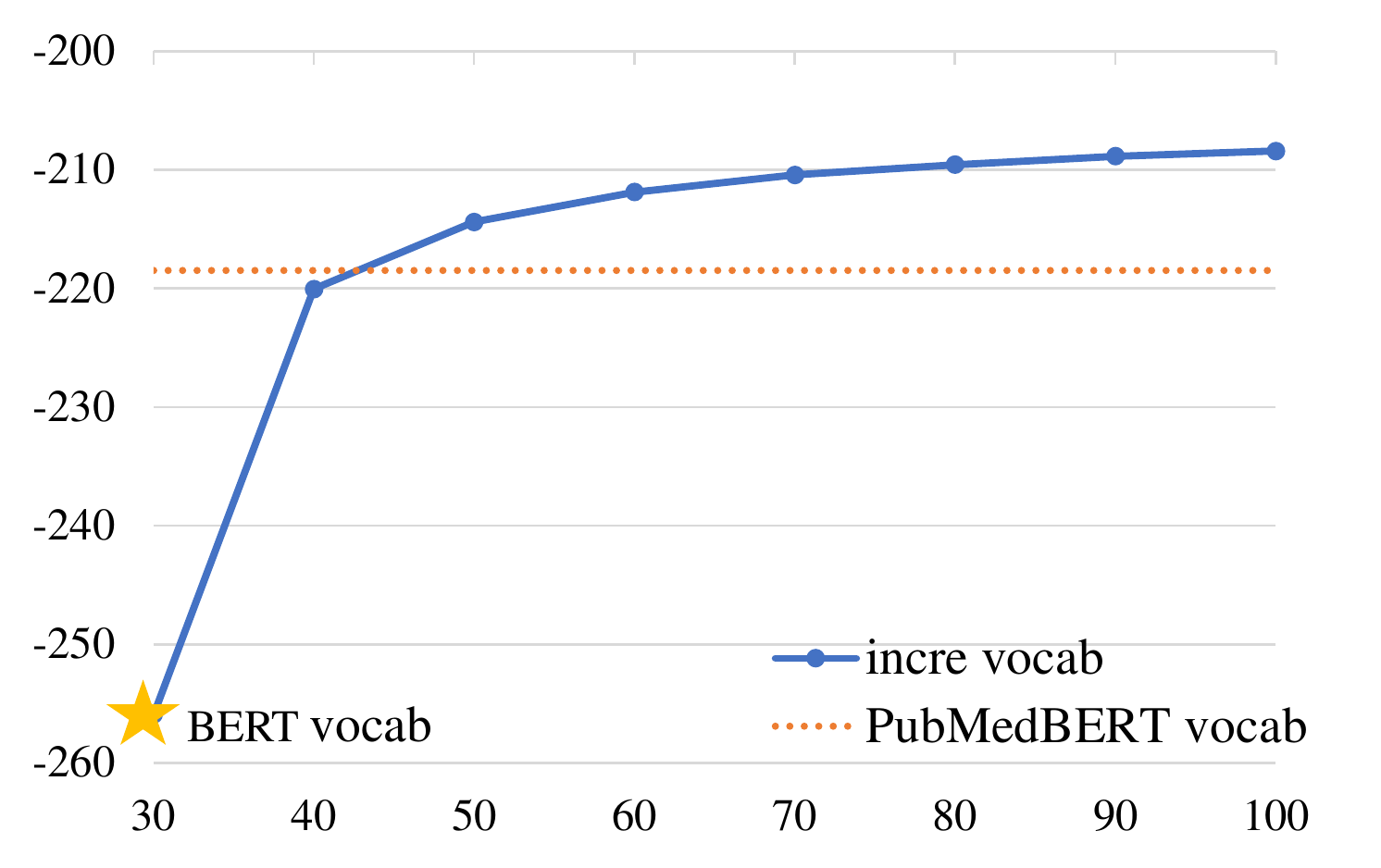}
   \caption{The $P(D)$ of different vocab sizes under biomedical domain. We use the BERT's vocabulary as the 30k vocabulary without vocabulary expanding. The PubMedBERT vocabulary is also 30k.}
   \label{fig:vocab_size_copare}
\end{figure}

We sample 550k sentences from the PubMed corpus and compute the occurrence probability $P(D)$ with different vocabulary sizes. The results are shown in Figure~\ref{fig:vocab_size_copare}. We compare the occurrence probability with BERT and PubMedBERT vocabularies. We observe that $P(D)$ reveals a logarithmic trend with substantial increases at the beginning and little influence after vocabulary size of 70k in the biomedical domain. The PubMedBERT vocabulary performs similarly to the 40k size vocabulary. We present the occurrence probability of different vocabulary sizes in Appendix~\ref{sec:probability}.

We propose a simple method to decide the size of the incremental vocabulary. Assume the probability at the time step $i-1$ is $P_{i-1}(D)$ and at the time step $i$ is $P_{i}(D)$. If the rise $\frac{P_{i}(D) - P_{i-1}(D)}{P_{i-1}(D)}$ is lower than a threshold $\delta$,  we regard the vocabulary size at the time step $i$ as the final size. 

\begin{algorithm}  
  
  \caption{Vocabulary Expansion}  
  \label{algo}
  \KwIn{Original vocabulary $raw\_vocab$, domain corpora $D$, threshold $\delta$ and  vocabulary size step $V_\Delta$}  
  \KwOut{$vocab_{final}$}  
  $token\_count \leftarrow$ whitespace split from $D$\;
  $P_0 \leftarrow$ computed from $raw\_vocab$\;
  $V_0 \leftarrow |raw\_vocab|$\;
  \Do {
        $\frac{P_{i} - P_{i-1}}{P_{i-1}}>\delta$
    } {
        vocabulary size $V_i \leftarrow V_{i-1} + V_\Delta$\;
        $sub\_count \leftarrow$ split token to subwords\; 
        Sort $sub\_count$ by frequency\;
        $incr\_vocab \leftarrow $keep $(V_i - V_0)$ subwords\;
        $vocab_{i} \leftarrow$ $raw\_vocab + incr\_vocab$\; 
        $P_i \leftarrow$ computed from $vocab_i$
        
    }
  return $vocab_{final} \leftarrow vocab_{i}$ \;  
\end{algorithm}
We expand the domain-specific vocabulary with the process shown in Algorithm~\ref{algo}. We implement our vocabulary expansion algorithm referring to SubwordTextBuilder in tensor2tensor\footnote{https://github.com/tensorflow/tensor2tensor}. In experiments, we set the threshold $\delta$ as $1\%$ and vocabulary size step $V_\Delta$ as 10k. Finally, we obtain the expanded vocabulary size of biomedical as 60k and computer science domain as 50k.

%% file: sec/ExperimentalDetails.tex
\section{Experiment Details}
We conduct our experiments in two domains: biomedical and computer science.
\subsection{Datasets}
\paragraph{Domain corpus:}

For the biomedical domain, we collect a 16GB corpus from PubMed\footnote{https://pubmed.ncbi.nlm.nih.gov/} abstracts to adapt our model. We use the latest collection and pre-process the corpora with the same process as PubMedBERT (we omit any abstracts with less than 128 words to reduce noise.). 

For the computer science domain, we use the abstracts text from the arXiv\footnote{https://www.kaggle.com/Cornell-University/arxiv} Dataset. We select abstracts in  computer science categories, collecting 300M entries for the corpus.

\paragraph{Fine-tuning tasks:}
For the biomedical domain, we choose three tasks: named entity recognition (NER), evidence-based medical information extraction (PICO), and relation extraction (RE). We perform entity-level F1 in NER task and word-level macro-F1 in the PICO task. The RE task uses the micro-F1 of positive classes evaluation. JNLPBA \citep{collier-kim-2004-introduction} NER dataset contains 6,892 disease mentions, which are mapped to 790 unique disease concepts with BIO tagging \cite{ramshaw1999text}. EBM PICO~\citep{nye2018corpus} datasets annotates text spans with four tags: Participants, Intervention, Comparator and Outcome. ChemProt~\cite{krallinger2017overview} dataset consists of five interactions between chemical and protein entities. We list the statistics of those tasks in Table~\ref{bio-task-info}.

We fine-tune two downstream tasks in the computer science domain. They are both classification tasks. The ACL-ARC~\cite{jurgens-etal-2018-measuring} dataset mainly focuses on analyzing how scientific works frame their contributions through different types of citations. SCIERC~\cite{luan-etal-2018-multi} dataset includes annotations for scientific entities, their relations, and coreference clusters. The statistics are available in Table~\ref{cs-task-info}.



\begin{table}[h]
\centering
\begin{tabular}{lrrr}
\bottomrule
\textbf{Dataset} & \textbf{Train} & \textbf{Dev} &\textbf{Test} \\
\hline
JNLPBA  & 46,750 & 4551 & 8,662 \\
EBM PICO & 339,167& 85321 & 16,364  \\
ChemProt & 18,035 & 11268 & 15,745 \\
\bottomrule
\end{tabular}
\caption{Biomedical dataset used in our experiment. All selected from BLURB\protect\footnotemark}
\label{bio-task-info}

\end{table}

\begin{table}[ht]
\centering
\begin{tabular}{lcccc}
\bottomrule
\textbf{Dataset} & \textbf{Train} & \textbf{Task} &\textbf{Test}& \textbf{Classes}\\
\hline
ACL-ARC &  1,688 & 114 & 139 & 6 \\
SCIERC &  3,219 & 455 & 974  & 7\\
\bottomrule
\end{tabular}
\caption{Computer science dataset used in our experiment. We use the same train, development, and test splits as \citet{dontstoppretraining2020}}
\label{cs-task-info}

\end{table}
\footnotetext{https://microsoft.github.io/BLURB/}

\subsection{Implementation}
We use the uncased version of $\text{BERT}_{\text{BASE}}$ ($12$ layers, $768$ hidden size) as the large model and the $\text{MiniLM}$ ($6$ layers, $384$ hidden size) as the small model.

To adapt the large model, we set the batch size at 8192 and the training step at 30,000. The peak learning rate was set to 6e-4. To adapt the small model, we set the batch size as 256 and the training step as 200,000. The learning rate is set to 1e-4. The maximum length of the input sequence was 512 and the token masking probability was 15\% for both the large model and the small model.

We implement MiniLM to compress large models and follow the setting of MiniLM, where the batch size was set to 256 and peak learning rate as 4e-4. We set the training step as 200,000.

For biomedical tasks, we follow the setting of PubMedBERT~\cite{2020Domain} to fine-tune these three tasks. For computer science tasks, we use the same setting as \citet{dontstoppretraining2020}. The concrete parameters are shown in Appendix~\ref{sec:hyparameter}.

%% file: sec/Results.tex
\section{Results}

\begin{table*}[ht]
\centering
\scalebox{0.83}{
\begin{tabular}{clllccccc}
\bottomrule
\textbf{Config}  & \multicolumn{1}{c}{\textbf{Type}}     & \textbf{Model} &  \textbf{Teacher}     & \textbf{JNLPBA} & \textbf{PICO} & \textbf{Chemprot}  & \textbf{Average} \\ \hline
\multirow{4}{*}{$L$=12; $d$=786} & \multirow{4}{*}{Large model} & BERT$^\dagger$ &  - & 78.63  & 72.34  & 71.86 &  74.28 \\
                                  && BioBERT$^\dagger$ & - &79.35& 73.18 & 76.14 & 76.22 \\
                                  && PubMedBERT$^\dagger$  & - & \textbf{80.06} & 73.38  & 77.24 &  76.89\\
                                  && AdaLM$^\diamondsuit$ & - & 79.46 & \textbf{75.47} &\textbf{78.41} &\textbf{77.74} \\
\hline
 \multirow{11}{*}{$L$=6; $d$=384} & \multirow{1}{*}{Small model} & MiniLM &- & 77.44 & 71.69 & 68.08 & 72.40\\ \cmidrule{2 - 8}
&\multirow{3}{*}{From scratch}     & BERT vocab (a) & - &77.89 &72.97                                   & 70.21 &  73.69\\
                                  && PubMed vocab (b)  & - &77.82 &73.82 & 70.32 &  73.99\\
                                 & & AdaLM vocab (c) &- &77.80 &73.39 & 70.86 &  74.02\\
\cmidrule{2 - 8}
& \multirow{3}{*}{Distill-then-Adapt}& BERT vocab (d) &  - & 78.63 & 74.00 & 72.28 & \underline{74.97} \\
                                 & & PubMed vocab (e)  &  - & 78.36   & 73.91 & 71.33 & \underline{74.53}\\
                                   & & AdaLM vocab (f)  & - & 78.77   & 74.23 & \textbf{72.29} & \underline{75.10}\\ 
\cmidrule{2 - 8}
&\multirow{3}{*}{Adapt-then-Distill}& Random initial (g)&  BERT & 77.98  & 72.38& 68.86 & 73.07 \\ 
                          && Random initial (h)  & PubMedBERT & 78.78  & 74.20 & 70.89 & \underline{74.62}   \\ 
                               & & Random initial (i) &AdaLM$^\diamondsuit$      & 78.98  & 74.78 & 71.51 & \underline{75.09} \\  \cmidrule{2 - 8}
                               & Adapt-and-Distill & Model (f) initial (j)   & AdaLM$^\diamondsuit$  &  \textbf{79.04} & \textbf{74.91} & 72.06 &  \underline{\textbf{75.34}}  \\ \bottomrule
\end{tabular}
}
\caption{Comparison between different strategies on biomedical tasks. The AdaLM$^\diamondsuit$ means we just adapt the large model without distillation. Scores of the methods marked with $^\dagger$ are taken from~\cite{2020Domain}. Underlined data marks the small models whose performances surpass the BERT model's performance. $L$ and $d$ indicate the number of layers
and the hidden dimension of the model.}
\label{table:biomedical}
\end{table*}

\begin{table*}[ht]
\centering
\scalebox{0.92}{
\begin{tabular}{clllccc}
\bottomrule
\textbf{Config}   & \multicolumn{1}{c}{\textbf{Type}}   & \textbf{Model}  &  \textbf{Teacher}     & \textbf{ACL-ARC} & \textbf{SCIERC}  & \textbf{Average} \\ \hline
\multirow{2}{*}{$L$=12; $d$=786 } &\multirow{2}{*}{Large model} & BERT &  -  &  64.92 & 81.14 & 73.03 \\
                                 & & AdaLM$^\diamondsuit$  & - & \textbf{73.61}  & \textbf{81.91} &  \textbf{77.76} \\ 
\hline
\multirow{8}{*}{$L$=6; $d$=384} & Small model & MiniLM  & -  & 61.5 & 72.88 & 67.19 \\
\cmidrule{2 - 7}
&\multirow{2}{*}{From scratch}     & BERT vocab (a)&   - & 62.48 & 74.93 &  68.70 \\
                             &     & AdaLM vocab (b) &   - & 59.57 & 74.93 & 67.25  \\ 
\cmidrule{2 - 7}
&\multirow{2}{*}{Distill-then-Adapt} & BERT vocab (c) &   - & 65.75 & 79.13 & 72.44 \\ 
                              &    & AdaLM vocab (d)  &  - & 65.93 & \textbf{79.88} & 72.91   \\
\cmidrule{2 - 7} 
&\multirow{2}{*}{Adapt-then-Distill}    & Random initial (e)  & BERT  & 63.12  & 77.89  & 70.50 \\ 
                           &   & Random initial (f)  &  AdaLM$^\diamondsuit$ & 66.21  &  77.04  & 71.62  \\  \cmidrule{2 - 7}
                            &  Adapt-and-Distill  & Model (d) initial (g) &  AdaLM$^\diamondsuit$ &\textbf{68.74}  & 78.88 & \underline{\textbf{73.81}}  \\ \bottomrule
\end{tabular}
}
\caption{Comparison between different strategies on computer science tasks. The AdaLM$^\diamondsuit$ is the adapted large model without compressing. We report averages across five random seeds. Data marked with underlines are the results of small models which outperform the BERT model's. $L$ and $d$ indicate the number of layers
and the hidden dimension of the model.}
\label{table:cs}
\end{table*}

The results of the tasks are shown in the Table~\ref{table:biomedical} and \ref{table:cs}.
We structure our evaluation by stepping through each of our three findings: 

(1) Domain-specific vocabulary plays a significant role in domain-specific tasks and expanding vocabulary with the general vocabulary is better than just using domain-specific vocabulary.

We observe improved results via the expanded vocabulary with both the large and small models. For large model, AdaLM achieves the best results under each domain, where 77.74 on biomedical domain tasks, beating BioBERT and PubMedBERT and 77.76 on the computer science domain tasks. 

For small models, in the biomedical domain, whether we train from scratch or distill-then-adapt with small models, incremental vocabulary models always perform better than the general vocabulary or just the domain-specific vocabulary. (When distill-then-adapt with the PubMed vocabulary, we initialize the word embedding in the same way as mentioned in Section~\ref{sec:adalm}). In addition, with distill-then-adapt, the model (f) (75.10) can surpass the BERT model (74.28). 

In the computer science domain, distill-then-adapt models with incremental vocabulary also show great performance. Model (d) achieves a comparable result of 72.91 as BERT and outperforms BERT in the ACL-ARC datasets with 65.93 (+1.01 F1). We also observe that when training from scratch, the results of Model (b) with incremental vocabulary are lower (1.45 lower) than that of model (a). This may be because after vocabulary expansion, a from-scratch model needs to be pretrained with more unlabeled data. 

(2) Continual pretraining on domain-specific texts from general language models is better than pretraining from scratch.

\citet{2020Domain} finds that for domains with abundant unlabeled texts, pretraining language models from scratch outperforms continual pretraining of general-domain language models. However, in our experiments, we find that general-domains model can help our model to learn the target domain better. In the biomedical domain, we use MiniLM model to initialize the model (d), (e) and (f) in distill-then-adapt setting. No matter which vocabulary is used, continual pretraining on domain-specific texts from general language models is better than pretraining from scratch. For AdaLM vocabulary, the model (f) gets 75.10, outperforming the model (c) trained from scratch with the same vocabulary by 1.08. On the other hand, for domains that do not have enormous unlabeled texts such as the computer science domain in our experiments, continual pretraining also showed better results. With continual pretraining, model (d) achieves higher results exceeding both model (b) (+5.66 F1) and model (c) (+0.47 F1).

(3) Adapt-and-Distill is the best strategy to develop a task-agnostic domain-specific small pretrained model.

In the Adapt-then-Distill part, our findings supports evidence from previous observations~\citep{wang2020minilm} that a better teacher model leads to a better student model. Using AdaLM which performs best among large models as the teacher model can yield good results: 75.09 in the biomedical domain and 71.62 in the computer science, better than other domain-specific large models. Furthermore, we find that a better student model for initialization can also help to get a better small model. In the Adapt-and-Distill part, we adapt large and small models into specific domains separately and then compress the adapted large model as the teacher with the adapted small model as initialization.
In the biomedical domain, the model (j), initialized from model (i), achieves the best result of 75.34 among the small models. It also outperforms the BERT model (+1.06 F1). In the computer science domain, model (g), initialized by model (d), is the only small model that outperforms BERT (+0.78 F1).

%% file: sec/Analysis.tex
\section{Analysis}
\subsection{Inference Speed} 
We compare AdaLM's parameters' size and inference speed with the BERT model in the biomedical domain in Table~\ref{speed-compare}. 
\begin{table}[ht]
\centering
\scalebox{1}{
\begin{tabular}{llcc}
\bottomrule
\textbf{Type}  &\textbf{Model}  & \textbf{\#Params}  & \textbf{Speedup} \\
\hline
\multirow{3}{*}{Large} &BERT & 109M & $\times 1.0$  \\ 
                            &PubMedBERT & 109M & $\times 1.0$ \\
                            &AdaLM vocab & 132M & $\times 1.07$ \\
                            
\hline
\multirow{2}{*}{Small} & BERT vocab & 22M & $\times 5.0 $ \\
                            &AdaLM   & 34M & $\times 5.1$\\
\bottomrule
\end{tabular}
}
\caption{Comparison of model's parameter size and the inference speed. The inference speedup is computed by the classification task ChemProt and evaluated on a single NVIDIA P100 GPU.}
\label{speed-compare}
\end{table}

First we can find that the vocabulary expansion yields marginal improvements on the model's inference speed. We added about 20M parameters in the embedding weights in the large model using AdaLM vocabulary, but its inference speed is slightly faster than BERT and PubMedBERT. Since most domain-specific terms are shattered into fragmented subwords, the length of the token sequence we get by using the incremental vocabulary is shorter than the length of the sequence got by the original vocabulary, which reduces the computation load. We list the change of the sequence length of the downstream tasks in Appendix~\ref{appen:seqlen}. Meanwhile, in the embedding layers, the model just needs to map the sub-words' id to their dense representations, which is little affected by the parameters' size. The small model shows the same trend. 

In addition, the small model AdaLM shows great potential. Compared with the 12-layer model of 768 hidden dimensions, the 6-layer model of 384 hidden dimensions is 3.3x smaller and 5.1x faster in the model efficiency, while performing similarly to or even better than $\text{BERT}_\text{BASE}$. 

\subsection{Impact of Training Time}
Pre-training often demands lots of time. In this section, we examine the adapted model's performance as a function of training time. Here we use the biomedical domain since its unlabelled texts are abundant and compare the large domain-specific adapted model with BioBERT. For every 24 hrs of continual pre-training, we fine-tuned the adapted model on the downstream tasks. For comparison, we convert the training time of BioBERT to the time it may take with the same computing resource of this work (16 V100 GPUs).
 \begin{table}[h]
 \centering
 
\begin{tabular}{lcc}
\bottomrule
\textbf{Model}  & \textbf{Training Time}  & \textbf{Average} \\
\hline
\multirow{4}{*}{AdaLM} & 0 hrs & 74.25 \\
                  & 24 hrs &  76.80\\
                  & 48 hrs &  77.36\\
                  & 72 hrs &  77.74\\
\hline
    BERT       & 0 hrs &  74.28\\                  
\hline
    BioBERT       & 120 hrs &  76.22\\
\bottomrule
\end{tabular}
\caption{Results with different pre-training time. In the table, AdaLM is the adapted large model without compressing.}
\label{time_compare}
\end{table}

We list the results in Table~\ref{time_compare}, we denote the large adapted model as AdaLM in the table. AdaLM at 0 hrs means that we fine-tune the initialized model directly without any continual pre-training. We find that BERT is slightly better than 0hr AdaLM and after 24 hrs, AdaLM outperforms BioBERT, which demonstrates that domain-specific vocabulary is very critical for domain adaption of pre-trained model. Our experiments demonstrate promising results in the biomedical domain. Under constrained computation, AdaLM achieves better performance compared to BioBERT. More details can be found in Appendix~\ref{sec:time}

\subsection{Impact of Vocabulary Size}
\label{sec: impact-of-size}
To understand the impact of the vocabulary size, we conduct some experiments with different vocabulary sizes in the biomedical domain. We select the biomedical large AdaLM model and to reduce the computation load, we set the batch size as 256 and step as 250K in our ablation studies. We show performance of the model with different sizes in Table~\ref{perform_diff_vocab}.

\begin{table}[ht]
\centering
\scalebox{0.9}{
\begin{tabular}{lccccc}
\bottomrule
 & 40k & 50k & 60k & 70k & 80k \\
 \hline
JNLPBA  &   78.84          & \textbf{79.02} & 78.91         & 78.94  & 79.01          \\
\hline
PICO    &   \textbf{75.09} & 74.81          & 74.99         & 74.58  & 75.00             \\
\hline
ChemProt & 76.10           & 76.80           & \textbf{77.21}& 76.40   & 76.85          \\
\hline
Average  &   76.67         & 76.87        & \textbf{77.03} & 76.64 & 76.95         \\

\bottomrule
\end{tabular}
}
\caption{The performance of different vocabulary sizes}
\label{perform_diff_vocab}
\end{table}

We observe that the model of 60k achieves the best results in our ablation studies. The result is a bit surprising. Despite having a larger vocabulary, the 70k and 80k model does not show a stronger performance. A possible explanation for these results may be that a larger vocabulary set may contain some more complicated but less frequent words, which cannot be learnt well through continual pre-training. For example, the word \emph{ferrocytochrome} exists in 70k and 80k vocabularies but is split into (\emph{`ferrocy', `\#\#tochrom', `\#\#e'}) in the 60k vocabulary. In our sampled data (about 550k sentences),  `\emph{ferrocytochrome}' appears less than 100 times, while the subword `\emph{\#\#tochrom}' appears more than 10k times and `\emph{ferrocy}' appears more than 200 times. The representation of those rare words cannot be learnt well due to the sparsity problem.

\subsection{Vocabulary Visualization}
The main motivation for using an the expanded vocabulary set is to leverage domain knowledge better. Compared to PubMedBERT which just uses the domain-specific vocabulary and initializes the model randomly, the keep of the general vocabulary and the general language model's weights may help us make good use of the existing knowledge and word embedding. 

\begin{figure}[h]
   \centering
   \includegraphics[scale=0.5]{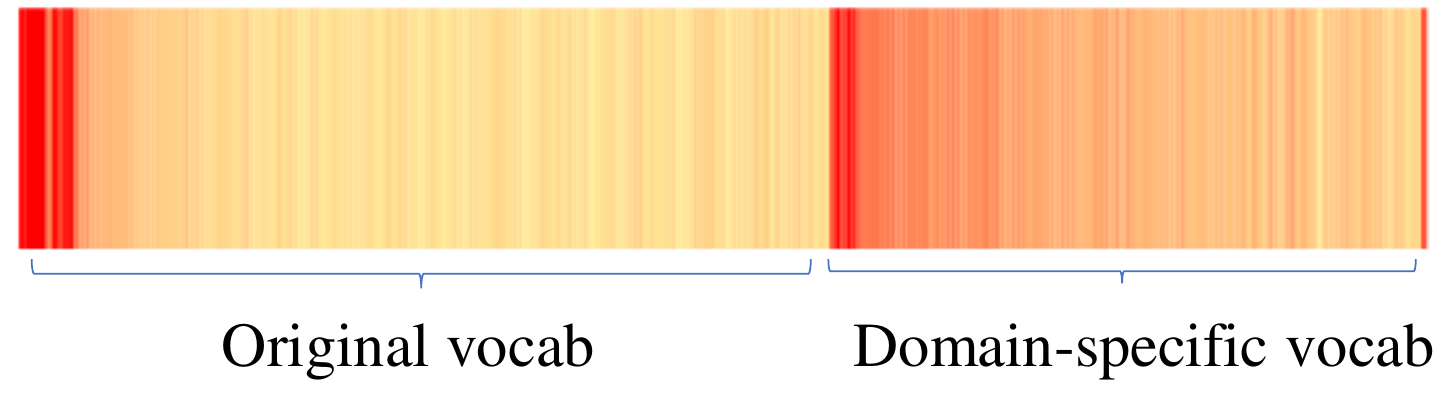}
   \caption{The $L2$-distance of the embedding layer. The deeper the color, the farther the distance. }
   \label{fig:vocab}
\end{figure}

To assess the importance of the expanded vocabulary, we compute the $L2$-distance of the embedding weights before and after pre-training in our AdaLM model in the biomedical domain in Figure~\ref{fig:vocab}.

We observe that the domain-specific vocabulary part changes a lot during the pre-training time, which indicates that our model learns much information about these domain-specific terms. We also observe that there is little change in many original sub-words’ embedding weights, which indicates that many general vocabularies can be used directly in continual training.



%% file: sec/Conclusion.tex
\section{Conclusion}
In this paper, we investigate several variations to compress general BERT models to specific domains. Our experiments reveal that the best strategy to obtain a task-agnostic domain-specific pretrained model is to adapt large and small models into specific domains separately and then compress the adapted large model with the adapted small model as initialization. We show that the adapted 6-layer model of 384 hidden dimensions outperforms the $\text{BERT}_{\text{BASE}}$ model while 3.3× smaller and 5.0× faster than $\text{BERT}_{\text{BASE}}$. Our findings suggest that domain-specific vocabulary and general-domain language model play vital roles in domain adaptation of a pretrained model. In the future, we will investigate more directions in domain adaptation, such as data selection and efficient adaptation.


%% file: sec/Appendix.tex
\appendix

\section{Occurrence probability of different vocabulary
sizes}
\label{sec:probability}

\begin{table}[ht]
\centering
\begin{tabular}{lc}
\hline
\textbf{Vocabulary} & \textbf{P(D)} \\
\hline
 \textbf{BERT} & -255.92\\
 \textbf{PubMed} & -218.49\\
\hline
 \textbf{40k vocab} & -220.06\\
 \textbf{50k vocab} & -214.40\\
 \textbf{60k vocab} & -211.88\\
 \textbf{70k vocab}& -210.44\\
 \textbf{80k vocab} & -209.57\\
 \textbf{90k vocab}& -208.86\\
 \textbf{100k vocab}& -208.42\\
 \hline
\end{tabular}
\caption{The $P(D)$ of different vocabulary under biomedical domain.}
\end{table}

\begin{table}[ht]
\centering
\begin{tabular}{lc}
\hline
\textbf{Vocabulary} & \textbf{P(D)} \\
\hline
 \textbf{BERT} & -211.14 \\
\hline
 \textbf{40k vocab} & -194.08 \\
 \textbf{50k vocab} & -192.56 \\
 \textbf{60k vocab} & -191.87 \\
 \textbf{70k vocab} & -191.45 \\
 \textbf{80k vocab} & -191.09 \\
 \textbf{90k vocab}& -190.76 \\
 \textbf{100k vocab}& -190.53 \\
 \hline
\end{tabular}
\caption{The $P(D)$ of different vocabulary under computer science domain.}
\end{table}

\section{Fine-tuning hyperparameters for downstream tasks}
\label{sec:hyparameter}

\begin{table}[ht]
\centering
\begin{tabular}{llcc}
\hline 
\textbf{Hyperameter} & \multicolumn{3}{c}{\textbf{Assignment}} \\
\hline
 & \textbf{NER} & \textbf{PICO} &\textbf{RE}\\
\hline
Batch size & 32 & \{16,32\} & 32 \\
Learning rate &\multicolumn{3}{c}{\{1e-5,3e-5,5e-5\}} \\
Epoch &\{30-40\} & \{10,15\} & \{40-50\}  \\
Dropout & \multicolumn{3}{c}{0.1}\\
\hline
\end{tabular}
\caption{Hyparameters we used to finetune on biomedical tasks.}
\label{bio:hyparameter-finetune}
\end{table}

\begin{table}[ht]
\centering
\begin{tabular}{llc}
\hline 
\textbf{Hyperameter} & \multicolumn{2}{c}{\textbf{Assignment}} \\
\hline
 & \textbf{ACL-ARC} &\textbf{SCIERC}\\
\hline
Batch size  &\multicolumn{2}{c}{16}\\
Learning rate&\multicolumn{2}{c}{2e-5} \\
Epoch  &\multicolumn{2}{c}{20}\\
Dropout & \multicolumn{2}{c}{0.1}\\
\hline
\end{tabular}
\caption{Hyparameters we used to finetune on computer science tasks.}
\label{cs: hyparameter-finetune}
\end{table}

\section{Sequence Length}
\label{appen:seqlen}
After the vocabulary expansion, the length of the token sequence may get shorter. We compute the average sentence length of the downstream tasks. We list the results in Table~\ref{append:lenght}  
\begin{table}[ht]
\begin{tabular}{lcc}
\bottomrule
\textbf{Dataset} & \textbf{Original Vocab} & \textbf{Incr. Vocab} \\ \midrule
ChemProt         & 66                      & 53                   \\
EBM PICO         & 36                      & 31                   \\
JNLPBA           & 41                      & 32                   \\ \midrule
ACL-ARC          & 53                      & 50                   \\
SCIERC           & 45                      & 42                  \\ \bottomrule
\end{tabular}
\caption{The sequence length tokenized by the original vocabulary and expanded vocabulary.}
\label{append:lenght}
\end{table}
\section{Results of different training time}
\label{sec:time}
We list the biomedical tasks' results of each pretraining time in the following table.
\begin{table}[ht]
\begin{tabular}{lcccc}
\hline
 & \textbf{0h} & \textbf{24h} & \textbf{48h} & \textbf{72h} \\
 \hline
 JNLPBA& 77.56 & 79.14 & 79.11 & 79.46 \\
 \hline
 PICO&73.29  & 74.22 & 75.28 & 75.36 \\
 \hline
 ChemProt& 71.91 & 77.06 & 77.69 & 78.42 \\
 \hline
 Average& 74.25 & 76.80 & 77.36 & 77.74 \\
 \hline
\end{tabular}
\caption{The performance of models with different pretraining time}
\end{table}



%% file: acl2021.bbl
\begin{thebibliography}{22}
\expandafter\ifx\csname natexlab\endcsname\relax\def\natexlab#1{#1}\fi

\bibitem[{Bostrom and Durrett(2020)}]{bostrom-durrett-2020-byte}
Kaj Bostrom and Greg Durrett. 2020.
\newblock \href {https://doi.org/10.18653/v1/2020.findings-emnlp.414} {Byte
  pair encoding is suboptimal for language model pretraining}.
\newblock In \emph{Findings of the Association for Computational Linguistics:
  EMNLP 2020}, pages 4617--4624, Online. Association for Computational
  Linguistics.

\bibitem[{Cheng et~al.(2017)Cheng, Wang, Zhou, and Zhang}]{cheng2017survey}
Yu~Cheng, Duo Wang, Pan Zhou, and Tao Zhang. 2017.
\newblock A survey of model compression and acceleration for deep neural
  networks.
\newblock \emph{arXiv preprint arXiv:1710.09282}.

\bibitem[{Collier and Kim(2004)}]{collier-kim-2004-introduction}
Nigel Collier and Jin-Dong Kim. 2004.
\newblock \href {https://www.aclweb.org/anthology/W04-1213} {Introduction to
  the bio-entity recognition task at {JNLPBA}}.
\newblock In \emph{Proceedings of the International Joint Workshop on Natural
  Language Processing in Biomedicine and its Applications
  ({NLPBA}/{B}io{NLP})}, pages 73--78, Geneva, Switzerland. COLING.

\bibitem[{Devlin et~al.(2019)Devlin, Chang, Lee, and
  Toutanova}]{devlin-etal-2019-bert}
Jacob Devlin, Ming-Wei Chang, Kenton Lee, and Kristina Toutanova. 2019.
\newblock \href {https://doi.org/10.18653/v1/N19-1423} {{BERT}: Pre-training of
  deep bidirectional transformers for language understanding}.
\newblock In \emph{Proceedings of the 2019 Conference of the North {A}merican
  Chapter of the Association for Computational Linguistics: Human Language
  Technologies, Volume 1 (Long and Short Papers)}, pages 4171--4186,
  Minneapolis, Minnesota. Association for Computational Linguistics.

\bibitem[{Dong et~al.(2019)Dong, Yang, Wang, Wei, Liu, Wang, Gao, Zhou, and
  Hon}]{dong2019unified}
Li~Dong, Nan Yang, Wenhui Wang, Furu Wei, Xiaodong Liu, Yu~Wang, Jianfeng Gao,
  Ming Zhou, and Hsiao{-}Wuen Hon. 2019.
\newblock \href
  {https://proceedings.neurips.cc/paper/2019/hash/c20bb2d9a50d5ac1f713f8b34d9aac5a-Abstract.html}
  {Unified language model pre-training for natural language understanding and
  generation}.
\newblock In \emph{Advances in Neural Information Processing Systems 32: Annual
  Conference on Neural Information Processing Systems 2019, NeurIPS 2019,
  December 8-14, 2019, Vancouver, BC, Canada}, pages 13042--13054.

\bibitem[{Gu et~al.(2020)Gu, Tinn, Cheng, Lucas, Usuyama, Liu, Naumann, Gao,
  and Poon}]{2020Domain}
Yu~Gu, Robert Tinn, Hao Cheng, Michael Lucas, Naoto Usuyama, Xiaodong Liu,
  Tristan Naumann, Jianfeng Gao, and Hoifung Poon. 2020.
\newblock Domain-specific language model pretraining for biomedical natural
  language processing.
\newblock \emph{arXiv preprint arXiv:2007.15779}.

\bibitem[{Gururangan et~al.(2020)Gururangan, Marasovi{\'c}, Swayamdipta, Lo,
  Beltagy, Downey, and Smith}]{dontstoppretraining2020}
Suchin Gururangan, Ana Marasovi{\'c}, Swabha Swayamdipta, Kyle Lo, Iz~Beltagy,
  Doug Downey, and Noah~A. Smith. 2020.
\newblock \href {https://doi.org/10.18653/v1/2020.acl-main.740} {Don{'}t stop
  pretraining: Adapt language models to domains and tasks}.
\newblock In \emph{Proceedings of the 58th Annual Meeting of the Association
  for Computational Linguistics}, pages 8342--8360, Online. Association for
  Computational Linguistics.

\bibitem[{Jiao et~al.(2020)Jiao, Yin, Shang, Jiang, Chen, Li, Wang, and
  Liu}]{jiao2019tinybert}
Xiaoqi Jiao, Yichun Yin, Lifeng Shang, Xin Jiang, Xiao Chen, Linlin Li, Fang
  Wang, and Qun Liu. 2020.
\newblock \href {https://doi.org/10.18653/v1/2020.findings-emnlp.372}
  {{T}iny{BERT}: Distilling {BERT} for natural language understanding}.
\newblock In \emph{Findings of the Association for Computational Linguistics:
  EMNLP 2020}, pages 4163--4174, Online. Association for Computational
  Linguistics.

\bibitem[{Jurgens et~al.(2018)Jurgens, Kumar, Hoover, McFarland, and
  Jurafsky}]{jurgens-etal-2018-measuring}
David Jurgens, Srijan Kumar, Raine Hoover, Dan McFarland, and Dan Jurafsky.
  2018.
\newblock \href {https://doi.org/10.1162/tacl_a_00028} {Measuring the evolution
  of a scientific field through citation frames}.
\newblock \emph{Transactions of the Association for Computational Linguistics},
  6:391--406.

\bibitem[{Krallinger et~al.(2017)Krallinger, Rabal, Akhondi, P{\'e}rez,
  Santamar{\'\i}a, Rodr{\'\i}guez et~al.}]{krallinger2017overview}
Martin Krallinger, Obdulia Rabal, Saber~A Akhondi, Mart{\i}n~P{\'e}rez
  P{\'e}rez, Jes{\'u}s Santamar{\'\i}a, GP~Rodr{\'\i}guez, et~al. 2017.
\newblock Overview of the biocreative vi chemical-protein interaction track.
\newblock In \emph{Proceedings of the sixth BioCreative challenge evaluation
  workshop}, volume~1, pages 141--146.

\bibitem[{Kudo(2018)}]{kudo2018subword}
Taku Kudo. 2018.
\newblock \href {https://doi.org/10.18653/v1/P18-1007} {Subword regularization:
  Improving neural network translation models with multiple subword
  candidates}.
\newblock In \emph{Proceedings of the 56th Annual Meeting of the Association
  for Computational Linguistics (Volume 1: Long Papers)}, pages 66--75,
  Melbourne, Australia. Association for Computational Linguistics.

\bibitem[{Lee et~al.(2020)Lee, Yoon, Kim, Kim, Kim, So, and
  Kang}]{lee2020biobert}
Jinhyuk Lee, Wonjin Yoon, Sungdong Kim, Donghyeon Kim, Sunkyu Kim, Chan~Ho So,
  and Jaewoo Kang. 2020.
\newblock Biobert: a pre-trained biomedical language representation model for
  biomedical text mining.
\newblock \emph{Bioinformatics}, 36(4):1234--1240.

\bibitem[{Liu et~al.(2019)Liu, Ott, Goyal, Du, Joshi, Chen, Levy, Lewis,
  Zettlemoyer, and Stoyanov}]{liu2019roberta}
Yinhan Liu, Myle Ott, Naman Goyal, Jingfei Du, Mandar Joshi, Danqi Chen, Omer
  Levy, Mike Lewis, Luke Zettlemoyer, and Veselin Stoyanov. 2019.
\newblock Roberta: A robustly optimized bert pretraining approach.
\newblock \emph{arXiv preprint arXiv:1907.11692}.

\bibitem[{Luan et~al.(2018)Luan, He, Ostendorf, and
  Hajishirzi}]{luan-etal-2018-multi}
Yi~Luan, Luheng He, Mari Ostendorf, and Hannaneh Hajishirzi. 2018.
\newblock \href {https://doi.org/10.18653/v1/D18-1360} {Multi-task
  identification of entities, relations, and coreference for scientific
  knowledge graph construction}.
\newblock In \emph{Proceedings of the 2018 Conference on Empirical Methods in
  Natural Language Processing}, pages 3219--3232, Brussels, Belgium.
  Association for Computational Linguistics.

\bibitem[{Nye et~al.(2018)Nye, Li, Patel, Yang, Marshall, Nenkova, and
  Wallace}]{nye2018corpus}
Benjamin Nye, Junyi~Jessy Li, Roma Patel, Yinfei Yang, Iain Marshall, Ani
  Nenkova, and Byron Wallace. 2018.
\newblock \href {https://doi.org/10.18653/v1/P18-1019} {A corpus with
  multi-level annotations of patients, interventions and outcomes to support
  language processing for medical literature}.
\newblock In \emph{Proceedings of the 56th Annual Meeting of the Association
  for Computational Linguistics (Volume 1: Long Papers)}, pages 197--207,
  Melbourne, Australia. Association for Computational Linguistics.

\bibitem[{Radford et~al.(2018)Radford, Narasimhan, Salimans, and
  Sutskever}]{radford2018improving}
Alec Radford, Karthik Narasimhan, Tim Salimans, and Ilya Sutskever. 2018.
\newblock Improving language understanding by generative pre-training.

\bibitem[{Ramshaw and Marcus(1995)}]{ramshaw1999text}
Lance Ramshaw and Mitch Marcus. 1995.
\newblock \href {https://www.aclweb.org/anthology/W95-0107} {Text chunking
  using transformation-based learning}.
\newblock In \emph{Third Workshop on Very Large Corpora}.

\bibitem[{Sanh et~al.(2019)Sanh, Debut, Chaumond, and
  Wolf}]{sanh2019distilbert}
Victor Sanh, Lysandre Debut, Julien Chaumond, and Thomas Wolf. 2019.
\newblock Distilbert, a distilled version of bert: smaller, faster, cheaper and
  lighter.
\newblock \emph{arXiv preprint arXiv:1910.01108}.

\bibitem[{Sun et~al.(2020)Sun, Yu, Song, Liu, Yang, and
  Zhou}]{sun2020mobilebert}
Zhiqing Sun, Hongkun Yu, Xiaodan Song, Renjie Liu, Yiming Yang, and Denny Zhou.
  2020.
\newblock \href {https://doi.org/10.18653/v1/2020.acl-main.195}
  {{M}obile{BERT}: a compact task-agnostic {BERT} for resource-limited
  devices}.
\newblock In \emph{Proceedings of the 58th Annual Meeting of the Association
  for Computational Linguistics}, pages 2158--2170, Online. Association for
  Computational Linguistics.

\bibitem[{Tai et~al.(2020)Tai, Kung, Dong, Comiter, and Kuo}]{tai2020exbert}
Wen Tai, HT~Kung, Xin~Luna Dong, Marcus Comiter, and Chang-Fu Kuo. 2020.
\newblock exbert: Extending pre-trained models with domain-specific vocabulary
  under constrained training resources.
\newblock In \emph{Proceedings of the 2020 Conference on Empirical Methods in
  Natural Language Processing: Findings}, pages 1433--1439.

\bibitem[{Wang et~al.(2020)Wang, Wei, Dong, Bao, Yang, and
  Zhou}]{wang2020minilm}
Wenhui Wang, Furu Wei, Li~Dong, Hangbo Bao, Nan Yang, and Ming Zhou. 2020.
\newblock \href
  {https://proceedings.neurips.cc/paper/2020/file/3f5ee243547dee91fbd053c1c4a845aa-Paper.pdf}
  {Minilm: Deep self-attention distillation for task-agnostic compression of
  pre-trained transformers}.
\newblock In \emph{Advances in Neural Information Processing Systems 33: Annual
  Conference on Neural Information Processing Systems 2020, NeurIPS 2020,
  December 6-12, 2020, virtual}.

\bibitem[{Zhang et~al.(2020)Zhang, Gangi~Reddy, Sultan, Castelli, Ferritto,
  Florian, Sarioglu~Kayi, Roukos, Sil, and Ward}]{zhang2020multi}
Rong Zhang, Revanth Gangi~Reddy, Md~Arafat Sultan, Vittorio Castelli, Anthony
  Ferritto, Radu Florian, Efsun Sarioglu~Kayi, Salim Roukos, Avi Sil, and Todd
  Ward. 2020.
\newblock \href {https://doi.org/10.18653/v1/2020.emnlp-main.440} {Multi-stage
  pre-training for low-resource domain adaptation}.
\newblock In \emph{Proceedings of the 2020 Conference on Empirical Methods in
  Natural Language Processing (EMNLP)}, pages 5461--5468, Online. Association
  for Computational Linguistics.

\end{thebibliography}
